\documentclass{article}

\usepackage{microtype}
\usepackage{graphicx}
\usepackage{subfigure}
\usepackage{booktabs} 

\usepackage{hyperref}

\usepackage[accepted]{icml2023}

\usepackage{amsmath}
\usepackage{amssymb}
\usepackage{mathtools}
\usepackage{amsthm}

\usepackage[capitalize,noabbrev]{cleveref}

\theoremstyle{plain}

\theoremstyle{definition}

\theoremstyle{remark}

\icmltitlerunning{Exploring the Versatility of Zero-Shot CLIP for Interstitial Lung Disease Classification}

\begin{document}

\twocolumn[
\icmltitle{Exploring the Versatility of Zero-Shot CLIP for Interstitial Lung Disease Classification}



\icmlsetsymbol{equal}{*}

\begin{icmlauthorlist}
\icmlauthor{Cara Van Uden}{stanford_cs,aimi}
\icmlauthor{Christian Bluethgen}{aimi}
\icmlauthor{Maayane Attias}{aimi,stanford_icme}
\icmlauthor{Malgorzata Polacin}{stanford}
\icmlauthor{Haiwei Henry Guo}{stanford_rad}
\icmlauthor{Neha Simha}{stanford_med_pulm}
\icmlauthor{Rishi Raj}{stanford_med_pulm}
\icmlauthor{Curtis Langlotz}{aimi}
\end{icmlauthorlist}

\icmlaffiliation{stanford_cs}{Department of Computer Science, Stanford University, Stanford, CA, USA}
\icmlaffiliation{aimi}{Center for Artificial Intelligence in Medicine and Imaging, Stanford University, Stanford, CA, USA}
\icmlaffiliation{stanford_icme}{Institute for Computational and Mathematical Engineering, Stanford University, Stanford, CA, USA}
\icmlaffiliation{stanford_rad}{Department of Radiology, Stanford University School of Medicine, Stanford, CA, USA}
\icmlaffiliation{stanford_med_pulm}{Division of Pulmonary, Allergy and Critical Care Medicine, Stanford University School of Medicine, Stanford, CA, USA}
\icmlaffiliation{stanford}{Stanford University, Stanford, CA, USA}

\icmlcorrespondingauthor{Cara Van Uden}{cvanuden@stanford.edu}

\icmlkeywords{Machine Learning, ICML}

\vskip 0.3in
]



\printAffiliationsAndNotice{}  

\begin{abstract}


Interstitial lung diseases (ILD) present diagnostic challenges due to their varied manifestations and overlapping imaging features. To address this, we propose a machine learning approach that utilizes CLIP, a multimodal (image and text) self-supervised model, for ILD classification. We extensively integrate zero-shot CLIP throughout our workflow, starting from the initial extraction of image patches from volumetric CT scans and proceeding to ILD classification using "patch montages". Furthermore, we investigate how domain adaptive pretraining (DAPT) CLIP with task-specific images (CT "patch montages" extracted with ILD-specific prompts for CLIP) and/or text (lung-specific sections of radiology reports) affects downstream ILD classification performance. By leveraging CLIP-extracted "patch montages" and DAPT, we achieve strong zero-shot ILD classification results, including an AUROC of 0.893, without the need for any labeled training data. This work highlights the versatility and potential of multimodal models like CLIP for medical image classification tasks where labeled data is scarce.
\end{abstract}

\section{Introduction}
\label{sec:introduction}
Interstitial lung diseases (ILD) form a diverse group of disorders that involve inflammation and fibrosis of the pulmonary tissue and that necessitate a consensus of radiologic, pathological, and clinical findings to establish an accurate diagnosis. Proper management of ILD requires comprehensive monitoring through CT scans and lung function tests to evaluate the evolution of the disease along with the effectiveness of treatment. As ILD encompasses a wide range of lung diseases with different etiologies and treatments, but often similar, overlapping imaging appearances, it can be challenging for clinicians to accurately identify disease progression and distinguish it from other causes of respiratory symptoms. There is currently no standardized approach for evaluating the disease progression in ILD patients, which can result in inconsistency of the interpretation of clinical data. This leads to a high inter-reader variability even among specialized radiologists and complicates the creation of sufficiently labeled datasets. 


To address these challenges, we propose a novel approach that leverages multimodal deep learning techniques, specifically focusing on the combination of lung CT scans and radiology reports, to improve ILD detection and classification. Our methodology harnesses CLIP (Contrastive Language-Image Pretraining), a multimodal self-supervised learning framework that has demonstrated remarkable capabilities in capturing rich semantic representations across different modalities.

The main contributions of our work can be summarized as follows:

\begin{enumerate}
    \item Task-specific patch retrieval: We perform zero-shot cross-modal retrieval by prompting CLIP to retrieve regions of interest ("patches") that are most likely to contain ILD-related abnormalities from each lung CT volume. This approach does not require any labeled examples or explicit annotations for ILD and demonstrates promising initial zero-shot classification performance.
    \item Task-specific domain-adaptive pretraining (DAPT): We explore a domain-adaptive pretraining (DAPT) strategy that fine-tunes the CLIP model using task-specific images (CLIP-extracted "patch montages" from CT) and text (lung-specific sections of the radiology report) and hypothesize that this may allow the model to adapt even further to the ILD domain. 
    \item Zero-shot ILD classification: We demonstrate strong zero-shot ILD classification performance by leveraging task-specific patch extraction and the multimodal representations learned through DAPT. Our best model achieves a zero-shot ILD classification AUROC of 0.893 without requiring any labeled examples during training. This capability significantly reduces the annotation burden, enabling effective ILD diagnosis even in scenarios where labeled data is scarce.
\end{enumerate}

By combining these components, our approach achieves strong performance in ILD detection and classification while leveraging a limited number of labeled examples. 

\section{Related Work}
\label{sec:related_work}

Machine learning approaches have been explored for interstitial lung disease (ILD) diagnosis, addressing various aspects such as pattern detection, quantification, and prognosis. Previous studies have focused on classifying ILD cases based on standard criteria, distinguishing between different ILD subtypes, and assessing disease progression. \cite{barnes2022machine,dack2023artificial} highlighted the importance of multimodal and longitudinal data for diagnostic and prognostic models for ILD. \cite{soffer2022artificial} highlighted the inter- and intra-observer variability and limited labeled data common in applications of machine learning to ILD. \cite{pawar2022two} present a two-stage hybrid approach for ILD classification using high-resolution computed tomography (HRCT) images, achieving accurate lung segmentation and classification into six ILD classes directly from HRCT, without the need for manual identification of the region of interest. In DeepILD \cite{wang2019weakly}, the authors combined semi-supervised K-means clustering and convolutional neural networks (CNNs) to perform ILD classification from CT images and achieved high performance across three chronic fibrosing ILD types. By integrating chest CT scans and clinical information, \cite{mei2023interstitial} leveraged multiple models to diagnose five types of ILD with multimodal data and determine a patient’s 3-year survival rate. Additionally, MIXTURE \cite{uegami2022mixture} leveraged deep learning models to extract pathologically significant findings in ILD based on expert pathologists' perspectives. These studies demonstrate the potential of machine learning in ILD diagnosis and prognosis. 

Multimodal models that integrate image and text modalities have shown significant advancements in medical imaging and text analysis. CLIP leverages large-scale pretraining on natural image-text pairs to learn powerful representations for tasks like zero-shot classification, cross-modal retrieval, and visual question answering (VQA) \cite{radford2021learning}. ConVIRT \cite{zhang2022contrastive}, GLoRIA \cite{huang2021gloria}, and CheXzero \cite{tiu2022expert}, apply similar approaches to CLIP for the medical domain, focusing on chest X-ray interpretation with datasets like MIMIC-CXR \cite{johnson2019mimic} or CheXpert \cite{irvin2019chexpert}. CheXzero specifically enables zero-shot multi-label classification without explicit manual annotations, achieving performance comparable to expert radiologists and surpassing previous label-efficient methods. Similarly, BioMed-CLIP \cite{zhang2023large}, pretrained on figure-caption pairs from biomedical research articles, achieves state-of-the-art results in various vision-language processing tasks. These multimodal models exemplify the potential of deep learning in medical imaging and text analysis, enabling improved performance and reducing the reliance on labor-intensive labeling efforts.

Our work addresses specific gaps in the current literature and is motivated by three key factors. Firstly, while previous multimodal research has primarily focused on 2D image data, such as chest X-rays, our study emphasizes the application of image/text approaches to volumetric (3D) CT data. Moreover, we specifically emphasize the utilization of fine-grained details in both image and text domains, using CT "patches" for images and lung-specific sections of text. Finally, our approach incorporates multimodal zero-shot methods at various stages of our experiment workflow, starting from the extraction of image patches from CT scans and progressing to classification, at the CT scan level, for ILD. By addressing these gaps, our work aims to enhance the understanding and classification of ILD in the context of multimodal analysis.

\section{Methods}
\label{sec:methods}

\subsection{Dataset}
\label{sec:dataset}

After obtaining IRB approval, a dataset of 2303 CT studies from 1321 patients was retrieved from the hospital's picture archiving and communication system (PACS). The cases were selected by querying the hospital's database for patients that had received at least one CT exam following a high-resolution CT (HRCT) scan protocol for the evaluation of ILD between. The corresponding text reports of all CT scans were manually reviewed by a board-certified radiologist to assess if findings indicative of ILD had been described. A primary diagnosis likely responsible for the lung appearance was determined by a review of radiology reports and clinical notes, following multi-disciplinary team (MDT) discussion notes if available. See Table \ref{tab:dataset} for a summary of the data.

Note that we create separate labeled validation ($D_\text{Val}$) and test sets ($D_\text{Test}$), even though we do not use any labeled data during model domain-adaptive pretraining (DAPT). $D_\text{Val}$ and $D_\text{Test}$ were created by randomly sampling a total of 100 and 100 patients from $D_\text{All}$, ensuring that half of them showed signs of ILD on CT. Two patients were removed from $D_\text{Test}$ after sampling because they had received lung transplants before the respective CT scan date. Before performing DAPT on $D_\text{All}$, we removed the test set from it. Therefore, the test set is disjoint from the pretrain and validation sets.

We use $D_\text{Val}$ to choose our initial patch montage configuration and CLIP model for DAPT, and we evaluate our final classification models after DAPT on $D_\text{Test}$. 

\begin{table}
    \centering
    \begin{tabular}{lccc}
        \toprule
         & $D_\text{All}$ & $D_\text{Val}$ & $D_\text{Test}$ \\
        \midrule
        Patients (n)  & 1321 & 100 & 98 \\
        CT Scans (n) & 2303 & 100 & 98 \\
        Age (y) & 63.8 & 62.8 & 64.4 \\ 
        Female (\%) & 57.5 & 52.0 & 58.2 \\
        ILD in CT (\%) & 86.4 & 50.0 & 51.0 \\
        Diagnosis (\%) & & & \\
        ~ IPF & 14.6 & 27.0 & 16.3 \\
        ~ SSc & 10.8 & 16.0 & 22.4 \\
        ~ HP & 11.3 & 12.0 & 17.3 \\
        ~ Other & 63.3 & 45.0 & 56.1 \\
        \bottomrule
    \end{tabular}
    \caption{Dataset statistics. CT: Computed tomography. HP: Hypersensitivity pneumonitis. ILD: Interstitial lung disease. IPF: Idiopathic pulmonary fibrosis. SSc: Systemic sclerosis.}
    \label{tab:dataset}
\end{table}

\subsection{Data preprocessing}
\label{sec:preprocessing}

\begin{figure*}[h]
    \centering
    \includegraphics[scale=0.65]{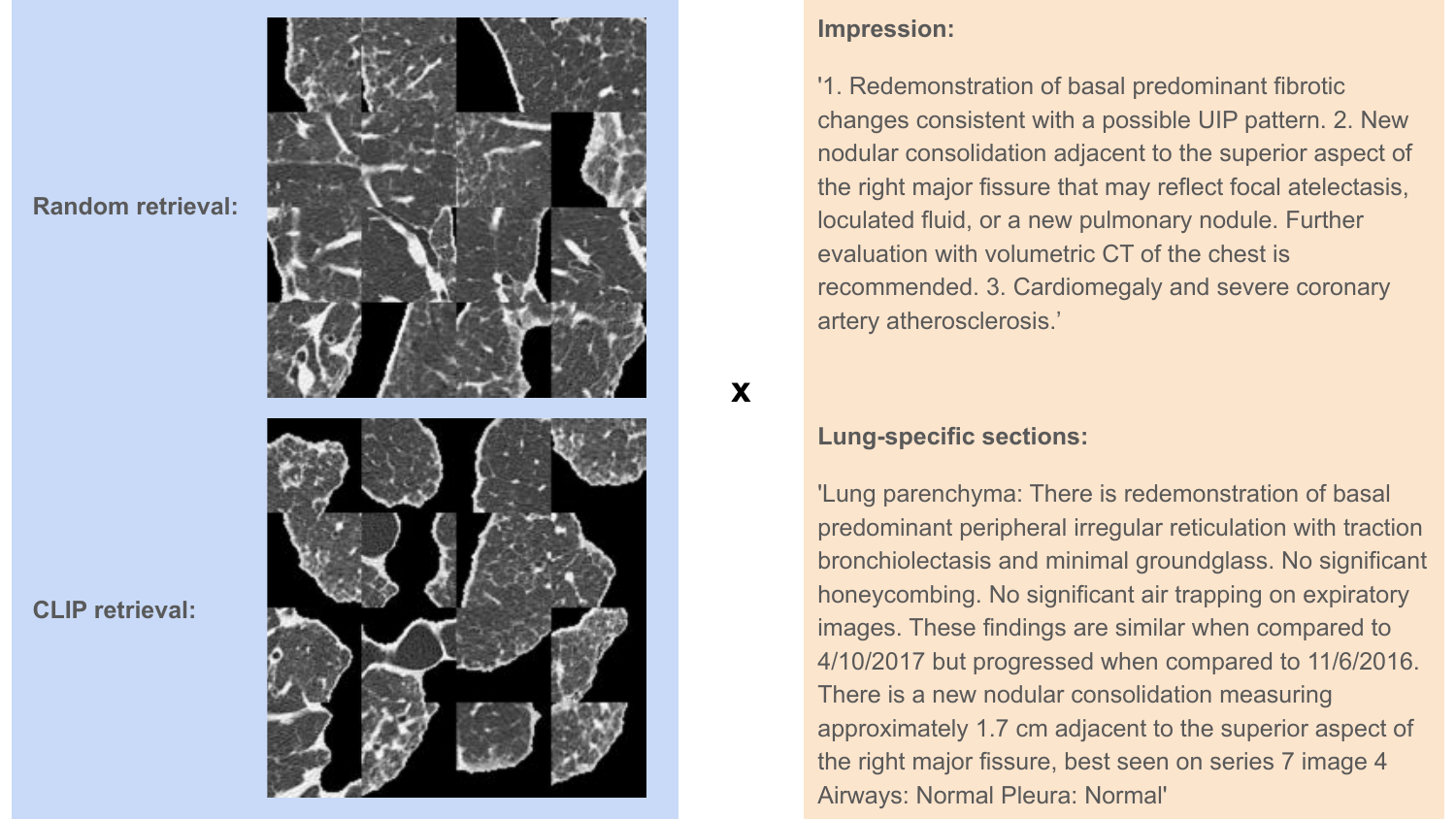}
    \caption{Example of $4 \times 4$ patch montage image and accompanying radiology report text. Patch montage images can be created with randomly-retrieved patches or with CLIP-retrieved patches, and text can be extracted from the impression or lung-specific sections.}
    \label{fig:montage_text_example}
\end{figure*}

We perform additional preprocessing on both the image (CT volume) and text (radiology report) data. See Figure \ref{fig:montage_text_example} for examples.

\begin{enumerate}
    \item Image:  We use only high-resolution CT (HRCT) scans. All HRCT scans were preprocessed to segment the lung tissue from the rest of the scan volume, using a combination of thresholding (-400 HU) and conventional erosion, opening and closing operations. From the segmented slices, we either extract image patches randomly or follow the patch-level classification process described in Section \ref{sec:zero_shot_methods}. We then combine the image patches into a "patch montage" image. We explore patch montage sizes of $4 \times 4$, $8 \times 8$, and $16 \times 16$. We also explore a "filter threshold", in which we enforce a minimum fraction of segmented lung included in the candidate image patch. We explore "filter thresholds" of 0.2, 0.5, and 0.8.
    \item Text: We extract either the impression (summary) section of the report or the lung-specific sections of the report. The lung-specific sections are those under the subheaders "lung parenchyma", "airways", and "pleura" in the radiology report.
\end{enumerate}

\subsection{Multimodal pretrained models}

We investigate three multimodal pretrained models:

\begin{enumerate}
    \item CLIP \cite{radford2021learning}: CLIP is a multimodal image-text model. This specific pretrained CLIP model uses a Vision Transformer (ViT) for its image encoder, and a Transformer as its text encoder. It was pretrained on the LAION-2B English subset of the LAION-5B dataset \cite{schuhmann2022laion}. We explore two different CLIP models with different ViT-B image encoders--ViT-B/16 and ViT-B/32.
    \item BioMed-CLIP \cite{zhang2023large}: BioMed-CLIP is a CLIP-style model pretrained on PMC-15M, a dataset of 15 million figure-caption pairs from biomedical research articles in PubMed Central. It consists of a PubMedBERT \cite{pubmedbert} text encoder and a Vision Transformer image encoder. With domain-specific adaptations, it excels in vision-language processing tasks like cross-modal retrieval, image classification, and visual question answering in the biomedical domain.
    \item CheXzero \cite{tiu2022expert}: CheXzero is a CLIP-style model initialized with pretrained weights from OpenAI's CLIP model. It utilizes the same ViT-B/32 architecture as our pretrained CLIP model. It is trained on the MIMIC-CXR dataset and uses the impressions section of each text report in its (image, text) pretraining pair. CheXzero demonstrates the ability to train a zero-shot model for domain-specific medical tasks. 
\end{enumerate}

\subsection{Domain-adaptive pretraining}
\label{sec:dapt_methods}

We perform domain-adaptive pretraining (DAPT) \cite{gururangan2020don} to adapt CLIP model to the ILD domain. We initialize from the CLIP model configuration that achieved the best zero-shot image-level ILD classification on the validation set. We then perform further contrastive pretraining using pairs of extracted "patch montages" (either random or CLIP-retrieved) and radiology report text (either impression or lung-specific sections). We train for up to 10 epochs with early stopping (patience of 1,000 steps) and save a model checkpoint every 100 steps. We perform a hyperparameter search over learning rate for five values in the range $[1e^{-5}, \cdots, 1e^{-3}]$. We hold out 10\% of the DAPT pretraining set as a validation set and choose the best model checkpoint as that which achieved the lowest validation set loss.

\subsection{Zero-shot classification}
\label{sec:zero_shot_methods}

To assess the model's zero-shot performance in patch and patch montage classification, we utilized a positive–negative softmax evaluation method for each disease, following the procedure in CheXzero \cite{tiu2022expert}. Here's how the procedure works: First, we calculate logits using positive prompts (e.g., "interstitial lung disease") and negative prompts ("no interstitial lung disease"). Then, we compute the softmax between the positive and negative logits. Finally, we consider the softmax probabilities of the positive logits as the likelihood of the disease being present in the CT image patch.

We perform two stages of zero-shot classification:

\begin{enumerate}
    \item Patch-level classification: We use this to rank the most task-specific (ILD-specific) patches from each CT volume when constructing "patch montages". The most ILD-specific patches are those with the highest softmax probability of "interstitial lung disease" over "no interstitial lung disease".
    \item Volume- ("patch montage-") level classification:  We use this to classify the "patch montage" extracted from each CT volume as positive or negative for ILD or one of its subtypes. The probability of disease is calculated as the softmax probability of "\{disease name\}" over "no \{disease name\}".
\end{enumerate}

\subsection{Reader study}
\label{sec:reader_study_methods}

We qualitatively evaluate our results via a reader study, in which we aim to compare randomly-retrieved, CLIP-retrieved, and DAPT-CLIP retrieved patch montages. We ask two board-certified radiologists to label and count the number of ILD-specific patches in each "patch montage" after patch extraction.

\section{Experiments}
\label{sec:experiments}

\begin{figure*}[h]
    \centering
    \includegraphics[scale=0.65]{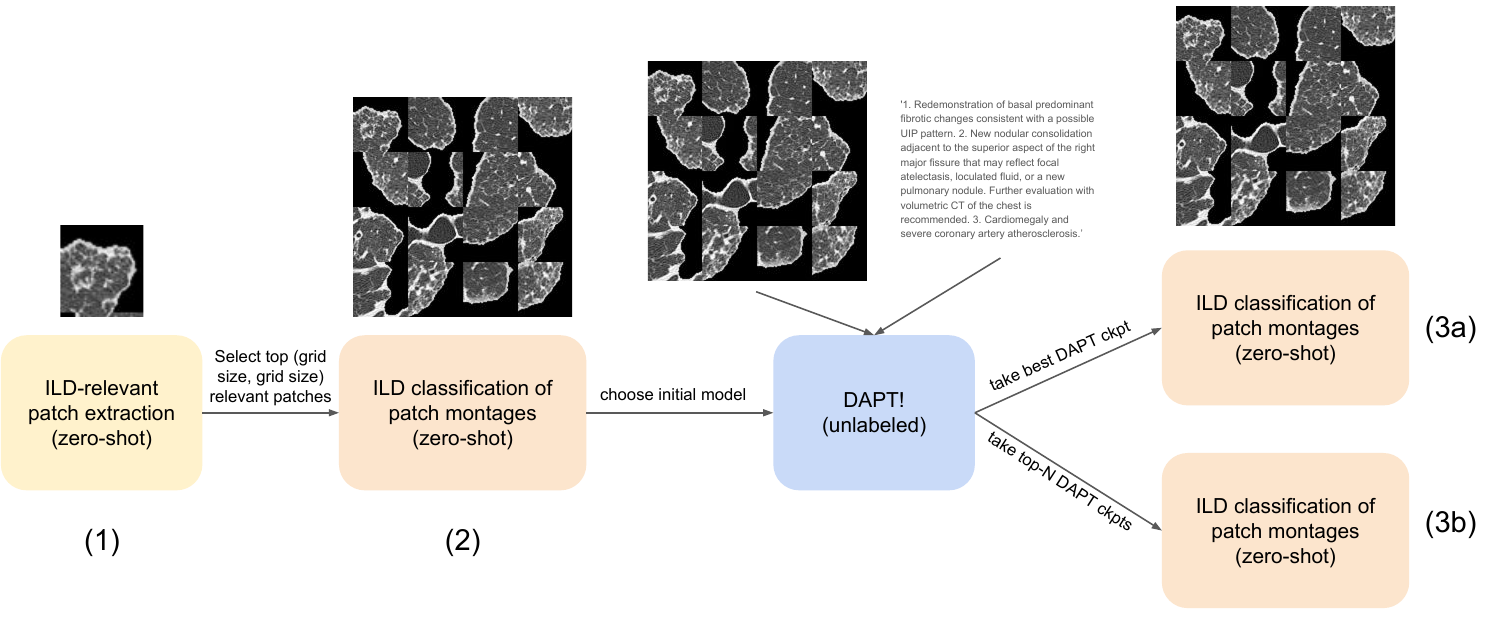}
    \caption{The experiment workflow. Note our usage of zero-shot CLIP at multiple stages of the workflow: (1) to choose our initial task-specific patches, (2) to perform initial zero-shot classification of "patch montages" and determine our model to use for DAPT, and (3a,b) to perform post-DAPT zero-shot classification using either a single model checkpoint or a model ensemble.}
    \label{fig:experiment_workflow}
\end{figure*}

Please see Figure \ref{fig:experiment_workflow} for an overview of our experiments. We perform the following experiments:
\begin{enumerate}
    \item Initial zero-shot classification of patch montages. Patch montages are extracted randomly or extracted via zero-shot prompting of a CLIP (CLIP, CheXzero, or BioMed-CLIP) pretrained model. Best performance here determines which CLIP model and montage size we use for DAPT.
    \item We perform DAPT for each combination of images (random; CLIP-retrieved extracted patch montage) and text (impression; lung section). This creates a set of four "DAPT-CLIP" models.
    \item Zero-shot classification of patch montages using DAPT-CLIP models. We evaluate using a single model checkpoint and using a model ensemble.
\end{enumerate}

\section{Results}
\label{sec:results}

\subsection{Initial zero-shot classification}
\label{sec:init_zero_shot}

We find that CheXzero, with randomly-retrieved patches, a patch montage of size $16 \times 16$, and a filter threshold of 0.5, achieves the best zero-shot ILD classification performance, achieving an AUROC of 0.771, AUPRC of 0.768, and F1-score of 0.730 on our validation set (see Table \ref{tab:init_zero_shot} for details). However, CheXzero, with either patch retrieval method, a patch montage of size $4 \times 4$, and a filter threshold of 0.5, achieve the next-best performance. As we test random \textit{and} CLIP (CheXzero)-retrieved performance further against DAPT-CLIP, we choose to use CheXzero and a $4 \times 4$ patch size moving forward. We therefore use this CheXzero $4 \times 4$ configuration to perform image preprocessing across the entire ILD dataset for later DAPT pretraining. 

For CheXzero with CheXzero-retrieved patches, a $4 \times 4$ patch montage size, and a 0.5 filter threshold, we achieve an AUROC of 0.674, AUPRC of 0.652, and F1-score of 0.705 on our validation set. In contrast to the relatively higher performance on the validation set, we later achieved an AUROC of 0.584, AUPRC of 0.602, and F1-score of 0.667 on the test set using this initial zero-shot approach. 


\begin{table*}[h]
    \centering
    \begin{tabular}{llc|ccc}
    \textbf{Patch Montage Method} & \textbf{Classification Model} & \textbf{Montage Size} & \textbf{AUROC} & \textbf{AUPRC} & \textbf{F1}    \\ \hline
                                    &                               & $4 \times 4$          & 0.470 &         0.476 &      0.658          \\
    Random                          & CLIP (ViT-B/16)               & $8 \times 8$          & 0.515 &         0.504 &      0.662          \\
                                    &                               & $16 \times 16$        & 0.432 &         0.475 &      0.667          \\ \hline
                                    &                               & $4 \times 4$          & 0.537 &         0.518 &      0.662          \\
    Random                          & CLIP (ViT-B/32)               & $8 \times 8$          & 0.485 &         0.491 &      0.662          \\
                                    &                               & $16 \times 16$        & 0.458 &         0.531 &      0.662          \\ \hline
                                    &                               & $4 \times 4$          & 0.541 &         0.532 &      0.662          \\
    Random                          & BioMed-CLIP (ViT-B/16)        & $8 \times 8$          & 0.512 &         0.507 &      0.676          \\
                                    &                               & $16 \times 16$        & 0.580 &         0.669 &      0.667          \\ \hline
                                    &                               & $4 \times 4$          & \underline{0.682} &         \underline{0.727} &      \underline{0.672}          \\
    Random                          & CheXzero (ViT-B/32)           & $8 \times 8$          & 0.624 &         0.639 &      0.656    \\
                                    &                               & $16 \times 16$        & \textbf{0.771} &         \textbf{0.768} &      \textbf{0.730}          \\ \hline \hline
                                    &                               & $4 \times 4$          & 0.422 &         0.439 &      0.681          \\
    CLIP (ViT-B/16)                 & CLIP (ViT-B/16)               & $8 \times 8$          & 0.502 &         0.513 &      0.671          \\
                                    &                               & $16 \times 16$        & 0.419 &         0.442 &      0.658          \\ \hline
                                    &                               & $4 \times 4$          & 0.405 &         0.461 &      0.662          \\
    CLIP (ViT-B/32)                 & CLIP (ViT-B/32)               & $8 \times 8$          & 0.422 &         0.457 &      0.662 \\
                                    &                               & $16 \times 16$        & 0.440 &         0.511 &      0.658          \\ \hline
                                    &                               & $4 \times 4$          & 0.617 &         \underline{0.635} &      \underline{0.686}          \\
    BioMed-CLIP (ViT-B/16)          & BioMed-CLIP (ViT-B/16)        & $8 \times 8$          & 0.566 &         0.544 &      0.672          \\
                                    &                               & $16 \times 16$        & 0.569 &         0.578 &      0.662          \\ \hline
                                    &                               & $4 \times 4$          & \textbf{0.674} &         \textbf{0.652} &      \textbf{0.705}          \\
    CheXzero (ViT-B/32)             & CheXzero (ViT-B/32)           & $8 \times 8$          & 0.530 &         0.515 &      0.653          \\
                                    &                               & $16 \times 16$        & \underline{0.650} &         0.612 &      0.682         
    \end{tabular}
    \caption{Zero-shot ILD classification performance on the validation set using patch montages. We vary the pretrained CLIP model used for patch retrieval and classification, as well as the patch montage size. We fix the filter threshold at 0.5; please see the complete results for filter thresholds 0.2 and 0.8 in Tables \ref{tab:init_zero_shot_0.2} and \ref{tab:init_zero_shot_0.8} in the Appendix. We find that CheXzero, with random patches and with a patch montage of size $16 \times 16$ achieves the best zero-shot ILD classification performance. However, CheXzero, with either patch type and with a patch montage of size $4 \times 4$ achieve the next-best performance. As we test random vs CLIP-retrieved performance further after DAPT, we choose to use CheXzero and a $4 \times 4$ patch size moving forward. Please see Section \ref{sec:init_zero_shot} for the best method's result on the test set.}
    \label{tab:init_zero_shot}
\end{table*}

\begin{figure*}[h]
    \centering
    \includegraphics[scale=0.65]{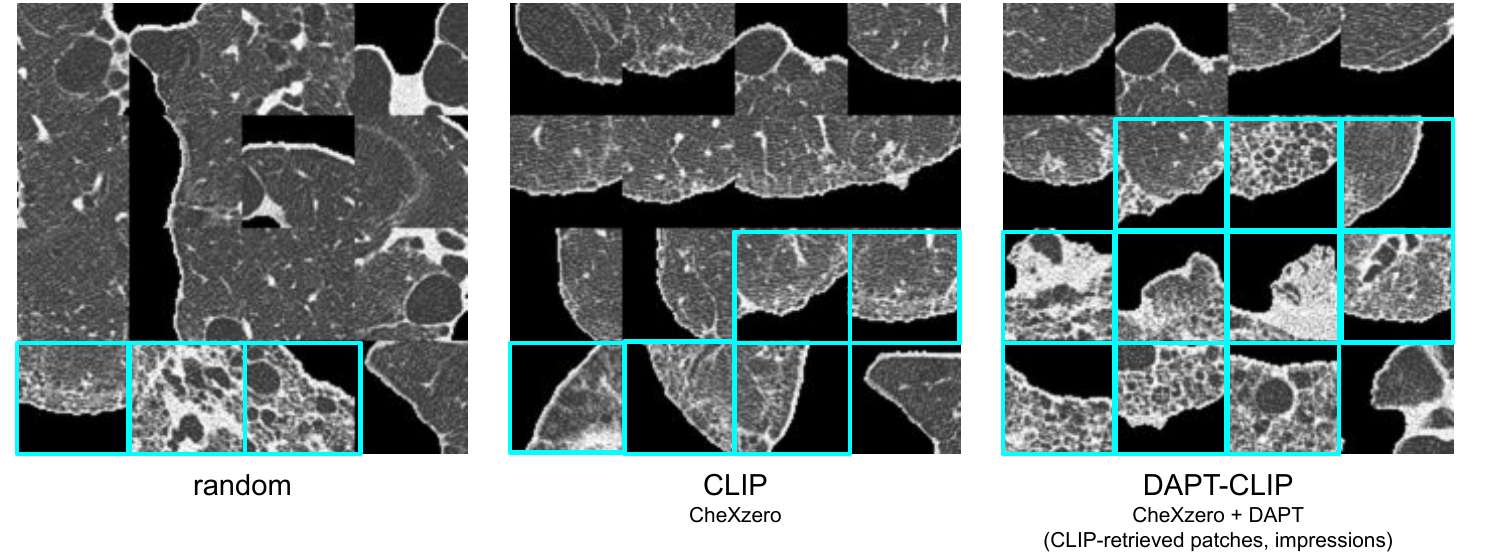}
    \caption{Examples of the patches retrieved randomly, by CLIP (CheXzero), and by DAPT-CLIP (our best CheXzero + DAPT model) for a single example CT scan. ILD-specific patches (boxed in blue) were labeled by a board-certified radiologist. For this CT scan, out of the 16 patches in the montage, random retrieves 3 relevant patches, CLIP retrieves 5, and DAPT-CLIP retrieves 10 patches.}
    \label{fig:retrieved_patches_example}
\end{figure*}

\subsection{DAPT zero-shot classification}
\label{sec:dapt_zero_shot}

We find that a DAPT-CLIP model pretrained on (CLIP (CheXzero)-retrieved $4 \times 4$ patch montage, impression) pairs achieves best performance when paired with DAPT-CLIP patch montage extraction at test time, achieving an AUROC of 0.893, AUPRC of 0.917, and F1-score of 0.824 on our test set. This is an increase of 0.309 AUROC, 0.315 AUPRC, and 0.157 F1 over the comparable model before DAPT (see Section \ref{sec:init_zero_shot}). Comparing our best Impressions-CLIP-$4 \times 4$ model against its random-patch-montage counterpart (Impressions-Random-$4 \times 4$), we see an increase in downstream classification performance when using CLIP to create patch montages during DAPT. In line with our hypothesis, we also see an even further increase in performance when then using the trained DAPT-CLIP model to retrieve patches at test time (Impressions-CLIP-$4 \times 4$ for patch montage retrieval + Impressions-CLIP-$4 \times 4$ for classification). Contrary to our initial hypothesis, we see a degradation in performance when using lung-related radiology report sections as text during DAPT across our DAPT-CLIP models. 

We also evaluate the classification performance of an ensemble of five DAPT-CLIP models, selected as the top five best checkpoints during DAPT by validation loss, and find no significant change in performance after ensembling. See Tables \ref{tab:dapt_zero_shot} and \ref{tab:dapt_zero_shot_ensemble} for details. 

Finally, we evaluate zero-shot ILD \textit{subtype} classification for three ILD subtypes: idiopathic pulmonary fibrosis (IPF), systemic sclerosis (SSc), and hypersensitivity pneumonitis (HP) with our best DAPT-CLIP model. Zero-shot classification of these subtypes is a more difficult task than general ILD classification; we achieve a zero-shot performance of 0.800 AUROC for IPF, 0.649 for SSc, and 0.639 for HP. See Table \ref{tab:dapt_zero_shot_subtypes} for details.

\begin{table*}[h!]
    \centering
    \begin{tabular}{ll|lll}
    \textbf{Classification Model (DAPT-CLIP)} & \textbf{Patch Montage Method} & \textbf{AUROC} & \textbf{AUPRC} & \textbf{F1} \\ \hline
                                              & Random                                 & 0.797 &         0.824 &      0.744       \\
    Impressions-Random-$4 \times 4$           & CLIP                                   & 0.857 &         0.868 &      0.811       \\
                                              & DAPT-CLIP                              & 0.802 &         0.846 &      0.752       \\ \hline
                                              & Random                                 & 0.798 &         0.818 &      0.776       \\
    Sections-Random-$4 \times 4$              & CLIP                                   & 0.844 &         0.858 &      0.807       \\
                                              & DAPT-CLIP                              & 0.821 &         0.841 &      0.792       \\ \hline
                                              & Random                                 & 0.835 &         0.814 &      0.781       \\
    Impressions-CLIP-$4 \times 4$             & CLIP                                   & 0.850 &         0.872 &      0.774       \\
                                              & DAPT-CLIP                              & \textbf{0.893} &         \textbf{0.917} &      \textbf{0.824}       \\ \hline
                                              & Random                                 & 0.770 &         0.797 &      0.674       \\
    Sections-CLIP-$4 \times 4$                & CLIP                                   & 0.786 &         0.824 &      0.742       \\
                                              & DAPT-CLIP                              & 0.760 &         0.810 &      0.716    \\ 
    \end{tabular}
    \caption{Zero-shot ILD classification on the test set after domain-adaptive CLIP pretraining (DAPT-CLIP). We test each combination of DAPT-CLIP classification model and test-time patch extraction method. We find that a DAPT-CLIP model pretrained on (CLIP (CheXzero)-retrieved $4 \times 4$ patch montage, impression) pairs achieves best performance when paired with DAPT-CLIP patch montage extraction at test time.}
    \label{tab:dapt_zero_shot}
\end{table*}

\begin{table*}[h!]
    \centering
    \begin{tabular}{l|lll}
    \textbf{ILD Subtype} & \textbf{AUROC} & \textbf{AUPRC} & \textbf{F1} \\ \hline
    IPF & 0.800 &                       0.403 &                    0.500 \\
    SSc & 0.649 &                       0.329 &                    0.442 \\
    HP & 0.639 &                      0.242 &                   0.351 \\
    \end{tabular}
    \caption{Zero-shot ILD subtype classification on the test set with the best DAPT-CLIP model (Impressions-CLIP-$4 \times 4$). The three subtypes classified here are idiopathic pulmonary fibrosis (IPF), systemic sclerosis (SSc), and hypersensitivity pneumonitis (HP).}
    \label{tab:dapt_zero_shot_subtypes}
\end{table*}

\begin{table*}[h]
    \centering
    \begin{tabular}{ll|lll}
    \textbf{Classification Model (DAPT-CLIP)} & \textbf{Patch Montage Method} & \textbf{AUROC} & \textbf{AUPRC} & \textbf{F1} \\ \hline
                                              & Random                                 & 0.806 &         0.812 &      0.783       \\
    Impressions-Random-$4 \times 4$           & CLIP                                   & 0.832 &         0.850 &      0.777       \\
                                              & DAPT-CLIP                              & 0.841 &         0.872 &      0.816       \\ \hline
                                              & Random                                 & 0.782 &         0.819 &      0.739       \\
    Sections-Random-$4 \times 4$              & CLIP                                   & 0.816 &         0.838 &      0.767       \\
                                              & DAPT-CLIP                              & 0.763 &         0.793 &      0.736       \\ \hline
                                              & Random                                 & 0.796 &         0.795 &      0.789       \\
    Impressions-CLIP-$4 \times 4$             & CLIP                                   & 0.882 &         0.885 &      0.808       \\
                                              & DAPT-CLIP                              & \textbf{0.892} &         \textbf{0.913} &      \textbf{0.833}       \\ \hline
                                              & Random                                 & 0.783 &         0.829 &      0.739       \\
    Sections-CLIP-$4 \times 4$                & CLIP                                   & 0.840 &         0.874 &      0.771       \\
                                              & DAPT-CLIP                              & 0.835 &         0.882 &      0.773      
    \end{tabular}
    \caption{Zero-shot ILD classification on the test set, using an ensemble of 5 models, after domain-adaptive CLIP pretraining (DAPT-CLIP). We test each combination of DAPT-CLIP classification model and test-time patch extraction method.}
    \label{tab:dapt_zero_shot_ensemble}
\end{table*}

\subsection{Reader study}
\label{sec:reader_study}

\begin{table*}[h]
    \centering
    \begin{tabular}{cc|cc}
    \textbf{ILD} & \textbf{Patch Montage Method} & \textbf{\% of Patch Montage with ILD (Average)} & \textbf{ICC(3,1) Score} \\ \hline
                    & Random      & \textbf{8.3\%}          & 0.656      \\
    Negative        & CLIP        & 9.6\%         & 0.422  \\
                    & DAPT-CLIP   & 9.0\%         & \textbf{0.757}      \\ 
    \hline
                    & Random      & 37.1\%        & 0.612       \\
    Positive        & CLIP        & 49.4\%        & 0.700       \\
                    & DAPT-CLIP   & \textbf{63.2\%}         & \textbf{0.794}      \\ 
    \end{tabular}
    \caption{Reader study results. We compare randomly-retrieved, CLIP-retrieved, and DAPT-CLIP retrieved patch montages. Two board-certified radiologists counted the number of ILD-specific patches in each patch montage. We find that DAPT-CLIP retrieves the most ILD patches (on average) and has the highest interreader agreement.}
    \label{tab:rs_results}
\end{table*}

We examine the patches retrieved by our best DAPT-CLIP (see Table \ref{tab:dapt_zero_shot}) model against random patch retrieval and CLIP patch retrieval in a set of reader studies. Please see Figure \ref{fig:retrieved_patches_example} for an example of the ILD-specific patches retrieved randomly, by the CLIP model (CheXzero) before DAPT, and by the best CLIP model (CheXzero) after DAPT. 


Based on the results of our reader study, where readers reported the number of ILD-specific patches (see Table \ref{tab:rs_results}), we can draw the following conclusions:

\begin{itemize}
    \item For the ILD-positive patients, DAPT-CLIP retrieval outperformed the two other methods. Readers counted on average $10.1$ out of $16$ patches ($63.2\%$) for the DAPT-CLIP method,  $7.9$ out of $16$ ($49.4\%$) for the CLIP method and $5.9$ out of $16$ patches ($37.1\%$) for the random method. This indicates that the DAPT-CLIP model-based method is more effective in retrieving ILD-specific patches for positive patients.
    
    \item For ILD-negative patients, all three methods show a low percentage of reported abnormal patches, which aligns with the absence of visible ILD abnormalities in their CT scans. Readers counted, on average, $1.4$ out of $16$ patches ($9.0\%$) for the DAPT-CLIP method, $1.5$ out of $16$ ($9.6\%$) for the CLIP method and $1.3$ out of $16$ patches ($8.3\%$) for the random method. These findings indicate that the models, particularly DAPT-CLIP, detect subtle clinical abnormalities that are also identified by the readers. However, it is important to consider that these abnormalities may not have clinical significance, or there is a possibility of false positives among the reported patches.
    
    \item To verify inter-reader agreement, we computed the interclass correlation coefficient (ICC) \cite{bartko1966intraclass,koo2016guideline}, a statistical measure used to quantify the level of consistency between readers. As seen in \ref{tab:rs_results}, DAPT-CLIP achieves the highest ICC for both ILD-positive and -negative patients across the three patch retrieval methods, and the ICC scores shows good agreement for the ILD positive ($0.794$) and negative patients ($0.757$) for DAPT-CLIP. This slight difference in ICC scores between the positive and negative patients could be attributed to the subjectivity and complexity of interpretation of ILD abnormal patches. The absence of apparent abnormalities for ILD negative patients could lead to differences in interpretation among the readers, resulting in lower agreement. On the contrary, the ILD positive group might have exhibited more consistent and visually striking abnormalities, making them easier to identify and agree upon.
\end{itemize}

\section{Conclusion}
\label{sec:conclusion}

In this work, we demonstrate the versatility and effectiveness of zero-shot CLIP in the classification of interstitial lung disease (ILD). By leveraging zero-shot CLIP at various stages of our experiment workflow, from extracting initial patches to "patch montage" classification, we achieve strong ILD classification results (AUROC of 0.893, AUPRC of 0.917, and F1-score of 0.824) without the need for any labeled training data. The combination of task-specific image patch retrieval and domain-adaptive pretraining, both with CLIP, proves to be a powerful approach for ILD classification. 

Moreover, our findings indicate potential avenues for further improvement and exploration. Varying the prompts used for patch retrieval can serve as a valuable data augmentation strategy when creating "patch montages," potentially improving the robustness and generalizability of our model. We also see potential in using these models for other tasks relevant for ILD, such as predicting ILD prognosis. Another promising direction is to extend the use of our approach to finer-grained patch retrieval and classification. We performed initial experiments in zero-shot ILD subtype classification; further work in this direction would improve the clinical utility of our approach and provide more targeted insights for patient care and treatment planning. By targeting specific clinical observations for patch retrieval, such as reticulation, ground-glass opacities, or honeycombing, we could potentially further improve performance of our approach for diagnosis and prognosis of ILD and its subtypes.

This work highlights the significant contributions that can be made by leveraging multimodal pretrained models like CLIP in versatile ways for medical image classification, particularly in settings with limited labeled data. We hope that this work inspires further exploration and research in this direction.

\section{Acknowledgements}
\label{sec:acknowledgements}

This study is supported by Boehringer Ingelheim Pharmaceuticals, Inc. (BIPI) in collaboration with the Stanford Interstitial Lung Diseases program and the Stanford Center for Artificial Intelligence in Medicine and Imaging. CL is supported in part by the Medical Imaging and Data Resource Center (MIDRC), funded by the National Institute of Biomedical Imaging and Bioengineering (NIBIB) of the National Institutes of Health under contract 75N92020C00021.

\clearpage
\newpage
\bibliography{example_paper}
\bibliographystyle{icml2023}

\clearpage
\newpage
\appendix
\onecolumn
\section{Appendix}

\begin{table*}[h]
    \centering
    \begin{tabular}{llc|ccc}
    \textbf{Patch Montage Method} & \textbf{Classification Model} & \textbf{Montage Size} & \textbf{AUROC} & \textbf{AUPRC} & \textbf{F1}    \\ \hline
               &          CLIP (ViT-B/16) &          4 &         0.432 &         0.463 &      0.662 \\
              Random &          CLIP (ViT-B/16) &          8 &         0.380 &         0.433 &      0.662 \\
               &          CLIP (ViT-B/16) &         16 &         0.521 &         0.521 &      0.658 \\  \hline
               &          CLIP (ViT-B/32) &          4 &         0.454 &         0.476 &      0.662 \\
              Random &          CLIP (ViT-B/32) &          8 &         0.552 &         0.587 &      0.667 \\
               &          CLIP (ViT-B/32) &         16 &         0.472 &         0.524 &      0.667 \\  \hline
               &          BioMed-CLIP (ViT-B/16) &          4 &         0.498 &         0.530 &      0.662 \\
              Random &          BioMed-CLIP (ViT-B/16) &          8 &         0.610 &         0.592 &      0.676 \\
               &          BioMed-CLIP (ViT-B/16) &         16 &         0.598 &         0.656 &      0.658 \\  \hline
               &             CheXzero (ViT-B/32) &          4 &         0.581 &         0.636 &      0.671 \\
              Random &             CheXzero (ViT-B/32) &          8 &         0.568 &         0.556 &      0.667 \\
               &             CheXzero (ViT-B/32) &         16 &         0.598 &         0.602 &      0.658 \\  \hline \hline
                     &          CLIP (ViT-B/16) &          4 &         0.470 &         0.518 &      0.658 \\
              CLIP (ViT-B/16) &          CLIP (ViT-B/16) &          8 &         0.452 &         0.468 &      0.658 \\
                     &          CLIP (ViT-B/16) &         16 &         0.423 &         0.436 &      0.662 \\ \hline
                     &          CLIP (ViT-B/32) &          4 &         0.424 &         0.466 &      0.658 \\
              CLIP (ViT-B/32) &          CLIP (ViT-B/32) &          8 &         0.443 &         0.462 &      0.658 \\
                     &          CLIP (ViT-B/32) &         16 &         0.485 &         0.535 &      0.658 \\ \hline
                     &          BioMed-CLIP (ViT-B/16) &          4 &         0.551 &         0.607 &      0.658 \\
              BioMed-CLIP (ViT-B/16) &          BioMed-CLIP (ViT-B/16) &          8 &         0.440 &         0.450 &      0.658 \\
                     &          BioMed-CLIP (ViT-B/16) &         16 &         0.601 &         0.633 &      0.676 \\  \hline
                     &             CheXzero (ViT-B/32) &          4 &         0.671 &         0.708 &      0.640 \\
              CheXzero (ViT-B/32) &             CheXzero (ViT-B/32) &          8 &         0.618 &         0.585 &      0.667 \\
                     &             CheXzero (ViT-B/32) &         16 &         0.653 &         0.635 &      0.692 \\
    \end{tabular}
    \caption{Zero-shot ILD classification performance on the validation set using patch montages. We vary the pretrained CLIP model used for patch retrieval and classification, as well as the patch montage size. We fix the filter threshold at 0.2.}
    \label{tab:init_zero_shot_0.2}
\end{table*}

\begin{table*}[h]
    \centering
    \begin{tabular}{llc|ccc}
    \textbf{Patch Montage Method} & \textbf{Classification Model} & \textbf{Montage Size} & \textbf{AUROC} & \textbf{AUPRC} & \textbf{F1}    \\ \hline
               &          CLIP (ViT-B/16) &          4 &         0.524 &         0.506 &      0.658 \\
              Random &          CLIP (ViT-B/16) &          8 &         0.388 &         0.462 &      0.658 \\
               &          CLIP (ViT-B/16) &         16 &         0.556 &         0.576 &      0.667 \\ \hline
               &          CLIP (ViT-B/32) &          4 &         0.440 &         0.497 &      0.662 \\
              Random &          CLIP (ViT-B/32) &          8 &         0.459 &         0.478 &      0.658 \\
               &          CLIP (ViT-B/32) &         16 &         0.564 &         0.612 &      0.681 \\ \hline
               &          BioMed-CLIP (ViT-B/16) &          4 &         0.435 &         0.448 &      0.658 \\
              Random &          BioMed-CLIP (ViT-B/16) &          8 &         0.507 &         0.575 &      0.658 \\
               &          BioMed-CLIP (ViT-B/16) &         16 &         0.620 &         0.651 &      0.667 \\ \hline
               &             CheXzero (ViT-B/32) &          4 &         0.599 &         0.627 &      0.676 \\
              Random &             CheXzero (ViT-B/32) &          8 &         0.753 &         0.742 &      0.706 \\
               &             CheXzero (ViT-B/32) &         16 &         0.636 &         0.614 &      0.705 \\ \hline \hline
                     &          CLIP (ViT-B/16) &          4 &         0.530 &         0.541 &      0.662 \\
              CLIP (ViT-B/16)       &          CLIP (ViT-B/16) &          8 &         0.558 &         0.581 &      0.662 \\
                     &          CLIP (ViT-B/16) &         16 &         0.402 &         0.448 &      0.662 \\ \hline
                     &          CLIP (ViT-B/32) &          4 &         0.424 &         0.476 &      0.658 \\
              CLIP (ViT-B/32)       &          CLIP (ViT-B/32) &          8 &         0.494 &         0.505 &      0.671 \\
                     &          CLIP (ViT-B/32) &         16 &         0.492 &         0.523 &      0.667 \\ \hline
                     &          BioMed-CLIP (ViT-B/16) &          4 &         0.504 &         0.528 &      0.667 \\
              BioMed-CLIP (ViT-B/16)       &          BioMed-CLIP (ViT-B/16) &          8 &         0.636 &         0.680 &      0.667 \\
                     &          BioMed-CLIP (ViT-B/16) &         16 &         0.575 &         0.644 &      0.657 \\ \hline
                     &             CheXzero (ViT-B/32) &          4 &         0.593 &         0.575 &      0.667 \\
              CheXzero (ViT-B/32)       &             CheXzero (ViT-B/32) &          8 &         0.660 &         0.631 &      0.693 \\
                     &             CheXzero (ViT-B/32) &         16 &         0.536 &         0.530 &      0.662 \\      
    \end{tabular}
    \caption{Zero-shot ILD classification performance on the validation set using patch montages. We vary the pretrained CLIP model used for patch retrieval and classification, as well as the patch montage size. We fix the filter threshold at 0.8.}
    \label{tab:init_zero_shot_0.8}
\end{table*}

\end{document}